\documentclass[sigconf]{acmart}
\pdfoutput=1

\usepackage{amsmath,amsfonts}
\usepackage{algorithmic}
\usepackage{graphicx}
\usepackage{textcomp}
\usepackage{xcolor}
\usepackage{algorithmic}
\usepackage[ruled,linesnumbered,noend]{algorithm2e}
\usepackage{tablefootnote}

\newenvironment{packeditemize}{
\begin{list}{$\bullet$}{
\setlength{\labelwidth}{6pt}
\setlength{\itemsep}{0pt}
\setlength{\leftmargin}{\labelwidth}
\addtolength{\leftmargin}{\labelsep}
\setlength{\parindent}{0pt}
\setlength{\listparindent}{\parindent}
\setlength{\parsep}{0pt}
\setlength{\topsep}{3pt}}}{\end{list}}

\AtBeginDocument{%
  \providecommand\BibTeX{{%
    \normalfont B\kern-0.5em{\scshape i\kern-0.25em b}\kern-0.8em\TeX}}}

\copyrightyear{2024}
\acmYear{2024}
\setcopyright{rightsretained}
\acmConference[GLSVLSI '24]{Great Lakes Symposium on VLSI 2024}{June 12--14, 2024}{Clearwater, FL, USA}
\acmBooktitle{Great Lakes Symposium on VLSI 2024 (GLSVLSI '24), June 12--14, 2024, Clearwater, FL, USA}
\acmDOI{10.1145/3649476.3658768}
\acmISBN{979-8-4007-0605-9/24/06}

\begin{document}

\setlength{\abovecaptionskip}{0pt}
\setlength{\belowcaptionskip}{5pt}
\setlength{\textfloatsep}{0pt}
\setlength{\floatsep}{0pt}
\setlength{\dbltextfloatsep}{0pt}
\setlength{\belowdisplayskip}{5pt}
\setlength{\abovedisplayskip}{0pt}

\title{FedTR: Federated Learning Framework with Transfer Learning for Industrial Visual Inspection}

\author{Vikash Sathiamoorthy$^{1}$, Shuo Huai$^{1,2}$,  Hao Kong$^{1,2}$, Di Liu$^3$, Wendy Yong Yi Loy$^1$, Christian Makaya$^4$, Daren Ho$^{5}$, Ravi Subramaniam$^{4}$, Qian Lin$^4$, Weichen Liu$^{2*}$}
\affiliation{\institution{$^1$HP-NTU Digital Manufacturing Corporate Lab, Nanyang Technological University, Singapore}
\country{}}
\affiliation{\institution{$^2$School of Computer Science and Engineering, Nanyang Technological University, Singapore}
\country{}}
\affiliation{\institution{$^3$Department of Computer Science, Norwegian University of Science and Technology, Norway}
\country{}}
\affiliation{\institution{$^4$HP Inc., Palo Alto, California, USA; $^5$HP Singapore, Singapore}
\country{}}


\thanks{
$^{*}$Corresponding author: Weichen Liu (liu@ntu.edu.sg).\\This study is partially supported under the RIE2020 Industry Alignment Fund – Industry Collaboration Projects (IAF-ICP) Funding Initiative, as well as cash and in-kind contribution from the industry partner, HP Inc., through the HP-NTU Digital Manufacturing Corporate Lab (I1801E0028). This research is also partially supported by the Ministry of Education, Singapore, under its Academic Research Fund Tier 1 (RG94/23), and Nanyang Technological University, Singapore, under its NAP (M4082282/04INS000515C130).}
\renewcommand{\shortauthors}{Vikash and Shuo, et al.}
\renewcommand{\authors}{Vikash Sathiamoorthy, Shuo Huai,  Hao Kong, Di Liu, Wendy Yong Yi Loy, Christian Makaya, Daren Ho, Ravi Subramaniam, Qian Lin and Weichen Liu}
\begin{abstract}
Federated learning (FL) is a collaborative learning scheme to train deep learning models, where collaborating parties can consolidate their models without sharing local data with other parties, hence preserving data privacy. Nevertheless, when implementing FL in Industrial visual inspection (IVI), the constraints posed by limited data availability and the intricate nature of the inspection tasks significantly impact the performance of the resulting model.
This paper introduces FedTR, a novel FL framework incorporating transfer learning designed for Autonomous  IVI, focusing on the challenging task of identifying label defects through end-to-end text recognition. Transfer learning is a method that leverages the knowledge of a pre-trained model to adapt to a different dataset.
FedTR initially trains the model using a publicly available dataset, after which performs the essential federated learning process with model fine-tuning on the distributed and limited private data.
Extensive experiment results demonstrate the effectiveness and feasibility of FedTR on private ink cartridge datasets for label defect identification. FedTR achieves an end-to-end text recognition word-level accuracy of 95.5\% and 94.2\% on homogeneous and heterogeneous data respectively. Additionally, it attains performance levels that are on par with those achieved through centralized training.
\end{abstract}





\maketitle

\section{Introduction}
Quality control for manufacturing line inspection is crucial in assembly lines as production quality is a key driver of profitability and business growth \cite{cho2005relationship}. Industrial visual inspection (IVI) is traditionally performed by human operators, which is very tedious and costly. Meanwhile, human factors, particularly fatigue, account significantly for the inconsistency in quality deficits \cite{yung2020examining}.

The digital manufacturing industry has seen a growing adoption of computer vision solutions for cost reduction and enhanced quality inspection. Collaboration among plants offers a solution to gather diverse images by pooling training data onto a central server for cloud-based training. However, the data are yet dispersed over different plants under the protection of privacy restrictions such as  Personal Data Protection Act (PDPA) in Singapore \cite{chik2013singapore} and General Data Protection Regulation (GDPR) across Europe \cite{voigt2017eu}.

With the limited and distributed nature of data and constraints, federated learning (FL) \cite{mcmahan2017communication} is introduced to address these issues. Federated learning enables distributed learning across multiple data sources without data sharing. In this technique, the learning model is deployed to manufacturing plants that can train a machine-learning model and share the model updates with a server.

\begin{figure*}[t]
\centering
\includegraphics[width=\textwidth]{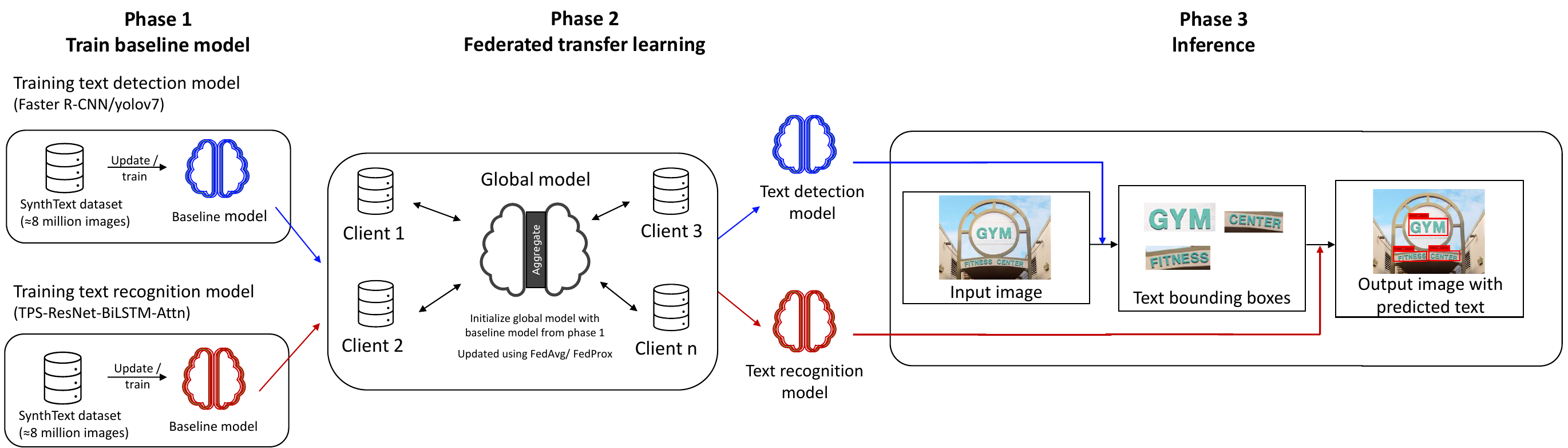}
\caption{FedTR framework}
\label{FedTR}
\end{figure*}
By eliminating the need for data transfer, FL guarantees dataset security and reduces communication costs. The plants that participate in FL can still operate in a country with strict laws on data export and intellectual property. Moreover, as model parameters are transferred instead of data, the communication overhead is reduced since the volume of data is typically much larger than the model parameters themselves. Nonetheless, FL prioritises privacy persevering over the accuracy of the learning model \cite{li2021survey}.

Therefore, to enhance the accuracy of federated learning, we propose to employ additional transfer learning techniques \cite{torrey2010transfer} to address the constraints imposed by limited data. Transfer learning can leverage the knowledge from other domains to train the model more efficiently. This is especially beneficial in the defect detection application as the available data is very limited. In summary, this paper makes the following main contributions:


\begin{packeditemize}
    \item We propose FedTR, an FL framework with transfer learning for IVI, which aggregates data from different manufacturing plants without compromising privacy and without extensive datasets.
    \item We show that our FedTR framework achieves comparable performance with the non-federated setting which includes individual and centralised training, in the context of end-to-end text recognition for both heterogeneous and non-heterogeneous datasets. 
    \item FedTR is extensible and can be a framework template for various applications, not limited to applications in the manufacturing industry. With the users’ privacy preserved and good model performance, FedTR can easily be deployed to other AI applications.
\end{packeditemize}


\vspace{-0.5cm}
\section{Related work}
\label{section:relatedwork}


\subsection{Federated learning}
Federated learning was first proposed by Google \cite{mcmahan2017communication} as a deep network learning from decentralised data by learning a shared model via aggregated locally computed updates. FL is a multi-step process.
Initially, clients are synchronized by the global model and trained using their data. Subsequently, clients transmit updates to the server, which aggregates them. This process iterates over multiple communication rounds. Crucially, decentralized data isn't shared, safeguarding user privacy. Thus, it's useful when users need to train models on larger datasets than their own but can't share the data itself. 
%
%
Nonetheless, FL encounters various challenges \cite{huai2022collate},
a fundamental challenge is heterogeneity, originating from two sources: system and statistical differences. System heterogeneity involves variations in hardware resources like memory, computation, and network. Statistical heterogeneity pertains to non-i.i.d. and unbalanced data \cite{huai2022collate}.

Generally, data heterogeneity degrades the performance of FL as it affects model convergence. In traditional FL, the local models are updated towards their local optima, which are often distant from each other due to non-i.i.d. data, the averaged global model may also deviate from the global optima \cite{li2021survey}. Li \textit{et al.} \cite{li2020federated} have shown that as data become more heterogeneous, convergence becomes worse for FedAvg \cite{mcmahan2017communication}, which is the FL algorithm used in the FedTR framework. FedTR framework mainly focuses on data heterogeneity in federated learning, but it can be integrated into system heterogeneity frameworks \cite{huai2022collate} for better utilizing the hardware resources.

\vspace{-0.5cm}
\subsection{Transfer learning}
Transfer learning \cite{torrey2010transfer} is inspired by the human ability to apply previously acquired knowledge to new and different tasks.The fundamental idea of transfer learning is to extract knowledge from one or more source tasks and apply this knowledge to a different target task. This process can significantly improve the learning efficiency and performance on the target task. Transfer learning is particularly useful in scenarios where the target task has limited data or resources, allowing the model to enhance its learning capabilities.

According to Zhuang\textit{ et al.} \cite{zhuang2020comprehensive}, transfer learning can be understood from two main perspectives: data-based and model-based. The data-based approach transfers knowledge by adjusting source domain data to better suit the target domain. This method modifies data to make it more applicable to the target domain. In contrast, the model-based approach involves techniques like model sharing, combining, and regularization. It focuses on adapting or merging existing models to facilitate learning in the target domain. In the model-based perspective, there are 4 main strategies: model control, parameter control, model ensemble and deep learning techniques. In the FedTR framework, fine-tuning, a commonly utilized technique in transfer learning, is categorized under parameter control.


\section{FedTR framework}
\label{section:methodology}
\subsection{Problem definition}
We have data from $N$ different users (manufacturing plants), denoted by $\{K_1, K_2, \ldots, K_N\}$, with image dataset $\{D_1, D_2, \ldots, D_N\}$, where the data may have different distribution.  In our problem, we want the users to collaborate to train a federated model, $M_{FL}$, where any user $K_i$ does not share their own data $D_i$. Suppose $M_{central}$ is the model trained using all dataset ${D_1, D_2, \ldots, D_N}$. This model is trained in a centralized manner, meaning all data subsets are aggregated and stored on a single node without any federated behaviors. 

The goals of our FedTR framework are to:
\begin{packeditemize}
    \item reduce the gap between $M_{central}$ and $M_{FL}$ performance
    \item ensure the performance of $M_{FL}$ is better  or same as each $M_i$.
\end{packeditemize}

Comparison is performed among individual training, centralized training, and federated training. In individual training, each user $K_i$ trains a model, $M_i$, using its own private data $D_i$. In centralised training, a model $M_{central}$ is trained by a concatenation of all the private data $D=D_1\cup D_2 \cup \ldots \cup D_N$.

\subsection{Autonomous End-to-end text recognition}
End-to-end text recognition involves text localization and text recognition. The text localization task involves identifying text in images to generate text bounding boxes while the text recognition task recognizes each character of the given text.

In the training process, the pipeline is divided into two stages: text detection and recognition, allowing for separate FL training in each stage. The text detection model uses the images as the input to learn the text bounding boxes while the recognition model uses the ground truth text bounding boxes as the input to learn the text itself. During the inference phase, the image is input to the text detection model to obtain word-level bounding boxes around the text in the image. Subsequently, each bounding box is forwarded to the text recognition model to generate the text prediction. Finally, the predicted texts and their corresponding bounding boxes are collated back to the original image.

\vspace{-0.4cm}
\subsection{FedTR framework overview}
\vspace{-0.1cm}

The primary goal of the FedTR framework is to support federated training for applications that involve limited private data.  The FedTR framework (see Fig. \ref{FedTR}) mainly consists of 3 phases. During training, as shown in Algorithm \ref{code:al1}, the initial model is first trained on a large open-source dataset in phase 1 (\textit{Line} 10). In phase 2, the model from phase 1 is distributed to each user for individual training on its private dataset to generate local models (\textit{Line} 1 - 6). The locally trained models from each user are then aggregated into a single model in the server (\textit{Line} 7 - 9) using the FedAvg algorithm  \cite{mcmahan2017communication}. This process is repeated for several communication rounds to obtain a trained global model (\textit{Line} 12 - 16). For end-to-end text recognition applications, the text detection and recognition models are trained independently, to support parallel training in phases 1 and 2. In the final phase, each user utilizes the two trained models to perform inference on their respective datasets, extracting bounding boxes and the corresponding predicted text from images.
In the FedTR framework, only a one-off training is required to generate a baseline model in phase 1. The resulting model can then be trained for different target applications in future phases. 

\vspace{-0.4cm}
\subsection{Federated learning}
\vspace{-0.1cm}

\begin{algorithm}[t]

\small
    \caption{FedTR}
    \label{code:al1}
    \KwIn{baseline model, training data, federated learning  parameters}
    \KwOut{updated global model}
      \DontPrintSemicolon
    \SetKwFunction{FMain}{LocalUpdate} 
    \SetKwFunction{Fa}{FedAvg}
    \SetKwProg{Fn}{Function}{:}{}
    \Fn{\FMain{$w$, $dataset$}}{
    $\mathcal{B} \gets$ split $dataset$ into batches of size $B$;\;
        \For{$e\gets1$ \KwTo $le$ }{
            \For {$b \in \mathcal{B}$ }{
            
                        $w \gets w - \eta\nabla l(w;b)$;\;
            }

        }
        return $w$;\;
    }

    \Fn{\Fa{$w_{1, 2,..., n}$}}{
        $w \gets \frac{1}{n} \sum_{i=1}^n w_i$;\;
        return $w$;\;
    }
Initialize baseline model and its parameters $w$;\;
$w^1 \gets$ Train $w$ with a public dataset;\;
\For{$r \gets 1$ \textbf{to} $R$}{

\For{$client_i$ \textbf{in} $clients_{all}$ \textbf{in parallel}}{

 $w_i^r \gets$ A copy of $w^r$;\;
  $w_i^{r+1} \gets$ LocalUpdate($w_i^r$, $dataset_i$);\;

}
$w^{r+1} \leftarrow$ FedAvg($w_{1,2,...,n}$);
}

\end{algorithm}

FedTR adopts FedAvg algorithm \cite{mcmahan2017communication} as the aggregation method in phase 2. FedAvg is an aggregation method that takes the average of the local models’ weights as the updated global model. In FedTR, the global model is initialized with the resulting model from phase 1 training to support transfer learning. 
At the start of each communication round, the server sends the global model to all users for local training. Each iteration of local training includes multiple local gradient steps on each user’s local data with stochastic gradient descent, followed by the model parameters averaging step through the server to get the new global model, which is then distributed to users for the next round of training.

\vspace{-0.3cm}
\subsection{Transfer learning}
The transfer learning technique used in FedTR is fine-tuning.
In FedTR, the pre-trained model from phase 1 serves as the initial model for the phase 2 federated training, where it is further trained with the target dataset. The entire model parameters are kept unfrozen for fine-tuning.

\section{Experiments}
\label{section:experiment}
\subsection{Dataset}
Three datasets are used for experiments: SynthText in the wild \cite{Gupta16}, custom ink cartridge generation I and generation II dataset. SynthText in the wild dataset\footnote{The SynthText dataset used was version 0.} consists of 858750 images with about 8 million synthetic word instances. The SynthText dataset is entirely used for baseline training with no data augmentation. 

The ink cartridge dataset consists of two generations. Generation I (Gen I) dataset has 377 images, with a standard size of (4208, 3120). The images contain the entire ink cartridge. While most of the images are taken from the top-down, a few images are taken from an angle. There are also some images with handwritten texts in which the handwritten texts are not annotated as ground truth. Generation II (Gen II) dataset contains 400 images of differing sizes. The images depict just the ink cartridge label itself and all the images are captured from the top down. Both cartridge gen I and II datasets contain defective and non-defective images. The ink cartridge dataset is divided into 3 subsets, with the train/val/test split ratio being 80/10/10. 

\vspace{-0.3cm}
\subsection{Implementation details}
Implementation of the FedTR framework is done using PyTorch and experiments are conducted to evaluate its effectiveness for end-to-end text recognition tasks. The experiments are tested on Nvidia Quadro GV100 GPU.

The target task is end-to-end text recognition. The text detection task is formulated as a one-class object detection problem. FedTR supports two text detection models: Faster R-CNN and Yolov7 \cite{wang2022yolov7} and one text recognition model: TPS-ResNet-BiLSTM-Attention  (TRBA). Faster R-CNN is initialised with pre-trained ResNet-50-FPN backbone while the other model parameters are randomly initialised. The Yolov7 model is initialized with pretrained weights from the official Yolov7 model implementation \cite{wang2022yolov7}. For the text recognition task, the objective is to recognise 68 case-insensitive alphanumeric and special characters. The TRBA, a 4-stage model is adapted from \cite{baek2019wrong} where the stages are transformation, feature extraction, sequence modelling and prediction.

To demonstrate the benefits and limitations of FedTR, a comparison is done between individual training (users train independently) and centralised training (all users gather their training data to collectively train the model).

\vspace{-0.3cm}
\subsection{Experimental setup}
\subsubsection{Baseline training with SynthText dataset}
The baseline training provides a trained model for use in the transfer learning phase of FedTR phase 2. The complete SynthText dataset was used for training purpose.
For text detection, Faster R-CNN and Yolov7 models were trained for 10 epochs for text detection while TRBA model was trained for 100 epochs for text recognition.

\subsubsection{Homogeneous data training: SynthText + Ink cartridge gen I dataset}

In this experiment, we simulate a scenario in which each of the manufacturing plants possesses data that is similar to that of the others. The data for the plants are considered to be independent and identically distributed (i.i.d.).

 During individual training, the model was trained locally on plant 1’s training dataset for 100 epochs. In federated learning, the model was trained for 20 communication rounds among 3 manufacturing plants. In each communication round, each plant locally trains with their respective dataset for 5 epochs. For text detection, the optimiser used is the stochastic gradient descent (SGD) update with a fixed learning rate of 0.1 and momentum of 0.5. For text recognition, the optimiser is Adadelta with a learning rate of 1.0, a decay rate of 0.95 and an epsilon of 1e-8. After the local training, the server aggregates the updates using the FedAvg algorithm. In centralised training, the model was trained locally on the entire cartridge gen I training dataset for 100 epochs.  All the models are initialised from the baseline model that had been trained on the SynthText dataset. The models are then evaluated on the centralised cartridge gen I validation dataset after training.

\subsubsection{Heterogeneous data training: SynthText + Ink cartridge gen I+II dataset}

In this experiment, We simulated a scenario in which various manufacturing plants employed different hardware setups for data collection, leading to data heterogeneity. Plant 1 has the cartridge gen I dataset while plant 2 has the cartridge gen II dataset. 

During individual training, each plant trains on its respective dataset for 100 epochs. Individual training of both plants is performed since the data of plants 1 and 2 are different. In federated learning, the model was trained for 20 communication rounds involving the two manufacturing plants. In each communication round, each plant locally trains with their respective dataset for 5 epochs. Experimental parameters are similar to homogeneous training for both tasks. After the local training, the server aggregates the updates using the FedAvg algorithm. In centralised training, the model was trained locally for 100 epochs on the dataset that includes both cartridge gen I and II training data. 

Similar to the previous experiment, all the models are initialised from the baseline model that was trained on the SynthText dataset. The resultant models are then evaluated on cartridge gen I and II validation datasets separately since each plant should not have access to the other plant’s data.

\subsection{Results and discussion}
\subsubsection{Homogeneous data training}

Table \ref{detection_homogeneous} shows the final validation metrics of different experiment settings for homogeneous data training. The individual, centralised and FedTR training all exhibit improvement from the baseline training. This outcome is expected as the aforementioned trainings are trained using  cartridge dataset and the models should learn to generalise to cartridge dataset.

The extent of improvement is different. FedTR achieves the highest validation F1 score of 0.726 when trained using the Yolov7 model. It is also observed that Yolov7 model has lower latency compared to Faster R-CNN model. The reason why FedTR performs better than individual training is that the individually trained model is exposed to its own limited data while FedTR model, indirectly is exposed to all data. FedTR having better performance motivates users (each manufacturing plant) to participate in federated learning since they benefit from enhanced model performance. 

It is also observed that FedTR performs better than centralised training. This is generally not expected, as the centralised training has direct access to all the training data as compared to FedTR, and hence should learn the features better. However, this outcome is attributed to high similarity in the dataset. Nevertheless, it's important to note that the difference in the F1 score is minimal.

For text recognition (see Table \ref{recognition_homogeneous}), all experimental settings exhibit improvement compared to the baseline. Individual training achieves the best final validation accuracy of 0.995. One potential explanation for this phenomenon is the similarity in data distribution between the SynthText and cartridge text images. 

Nevertheless, the difference between validation accuracy of experimental settings is small. Similar to text detection, FedTR exhibits a slightly lower but comparable accuracy of 0.991 when compared to centralised training, which achieves an accuracy of 0.994.

As explained earlier, the text detection and recognition models are evaluated separately. An end-to-end text recognition performance for FedTR was also evaluated on the cartridge gen I validation dataset. The input image is fed into the text detection model to produce predicted text bounding boxes, which are subsequently forwarded to the text recognition model to generate the predicted text. FedTR achieves \textbf{95.5\%} word-level accuracy. Upon visualisation, the texts are detected and recognised properly, with some instances of punctuation marks not detected and recognised.

\begin{table}[t]
\small
\centering
\caption{Text detection results for homogeneous data}
\label{detection_homogeneous}
\begin{tabular}{lll}
\hline \hline
Experiments & Training setting & F1 score \\
\hline 
Baseline (phase 1) & SynthText for 10 epochs  & 0.248    \\
Individual training & Subset of cartridge gen I   & 0.686    \\
& (100 images) for 100   epochs &  \\
Centralised training & Cartridge gen I for 100 epochs   & 0.701    \\
FedTR – frcnn & Cartridge gen I with 3 users & 0.718  \\
(phase 2)\tablefootnote{\label{FLconfig}Federated training was performed with 20 communication rounds and 5 local epochs for all experiments.} & &  \\
FedTR – yolov7 & Cartridge gen I with 3 users & \textbf{0.726}  \\
(phase 2)\footref{FLconfig} & &  \\
\hline \hline
\end{tabular}

\end{table}

\begin{table}[t]
\small
\centering
\caption{Text recognition results for homogeneous data}
\label{recognition_homogeneous}
\begin{tabular}{lll}
\hline \hline
Experiments & Training setting& Val. acc. \\
\hline 
Baseline (phase 1) & SynthText for 100 epochs  & 0.955   \\
Individual training & Subset of cartridge gen I  & \textbf{0.995}    \\
& (100 images) for 100   epochs &  \\
Centralised training & Cartridge gen I for 100 epochs    & 0.994    \\
FedTR – FedAvg & Cartridge gen I with 3 users & 0.991  \\
(phase 2) & &  \\
\hline \hline
\end{tabular}

\end{table}

\subsubsection{Heterogeneous data training}

From the text detection results (see Table \ref{detection_heterogeneous}), when each user performs individual training, its model only performs well on its own dataset. For instance, individual training for plant 2 achieves a 0.907 validation F1 score on its own validation dataset but only 0.147 for plant 1’s validation dataset. When comparing FedTR with individual training, even though the model performance of each user may decrease with federated learning, the FedTR model is more robust to data heterogeneity. The performance of FedTR surpasses that of individually training a model on one plant's dataset and then validating it on another plant's dataset, and vice versa.

Moreover, FedTR has comparable performance with centralised training. Although there is a slight drop in FedTR performance as compared to centralised training, the drop is acceptable as the training is done under a federated setting and the model does not have direct access to all the training data. 

For text recognition (see Table \ref{recognition_heterogeneous}), the individually trained models perform effectively on their respective datasets, similar to text detection result. In the context of text detection, individual training for both plants delivers the best results. However, for the specific datasets, centralised training excels in the case of plant 1, while FedTR outperforms for plant 2. However, the numerical difference (at most 0.7\% difference) of the trainings is very minimal.

The end-to-end text recognition performance of FedTR was assessed using the combined validation dataset (data from both plants). FedTR achieves \textbf{94.2\%} word-level accuracy. 

\begin{table}[t]
\small
\centering
\caption{Text detection results for heterogeneous data}
\label{detection_heterogeneous}
\begin{tabular}{llll}
\hline \hline
Experiments & Training setting & Plant 1 & Plant 2\\
& & val data & val data \\
& & F1 score & F1 score \\
\hline 
Baseline (phase 1) & SynthText for 10 epochs  & 0.253 & 0.428    \\
Individual training & Cartridge gen I for  & \textbf{0.692} & 0.472    \\
(plant 1) &  100 epochs & &  \\
Individual training & Cartridge gen II for  & 0.147 & \textbf{0.907}    \\
(plant 2) & 100 epochs & &  \\
Centralised training & Cartridge gen I+II for    & 0.686 & 0.715    \\
& 100 epochs & & \\
FedTR – frcnn & Cartridge gen I for plant 1, & 0.653 & 0.719  \\
(phase 2)& cartridge gen II for plant 2 & & \\
FedTR – yolov7 & Cartridge gen I for plant 1, & 0.651 & 0.711  \\
(phase 2)  & cartridge gen II for plant 2 & & \\
\hline \hline
\end{tabular}

\end{table}

\begin{table}[t]
\small
\centering
\caption{Text recognition results for heterogeneous data}
\label{recognition_heterogeneous}
\begin{tabular}{llll}
\hline \hline
Experiments & Training setting & Plant 1 & Plant 2\\
& & val acc & val acc \\
\hline 
Baseline (phase 1) & SynthText for 100 epochs  & 0.955 & 0.971   \\
Individual training & Cartridge gen I for & 0.993 & 0.957    \\
(plant 1) & 100 epochs & &  \\
Individual training & Cartridge gen II for  & 0.964 & 0.996    \\
(plant 2) & 100 epochs & &  \\
Centralised training & Cartridge gen I+II for  & \textbf{0.997} & 0.990    \\
& 100 epochs & & \\
FedTR – FedAvg & Cartridge gen I for plant 1, & 0.989 & \textbf{0.997}  \\
(phase 2) & cartridge gen II for plant 2 & & \\
\hline \hline
\end{tabular}

\end{table}

\section{Conclusion}
\label{section:conclusion}
In conclusion, the FedTR framework has been designed for the application of end-to-end autonomous text recognition as a practical use case. It leverages two main techniques: federated learning and transfer learning. While federated learning solves the problem of isolated data, transfer learning reduces the impact of limited data since model performance is usually dependent on a large amount of training data. FedTR framework can hence support multi-user collaboration while reducing the concern of having insufficient data to train a meaningful model, as well as addressing data privacy issues in cloud adoption. 

Experiments have demonstrated the effectiveness of the FedTR framework. First, users with limited data can benefit from FedTR in terms of better model performance, which provides the motivation for users to participate in federated learning. Furthermore, FedTR can attain performance levels that are on par with those achieved through centralized training. Overall, FedTR achieves 95.5\% and 94.2\% word-level accuracy for end-to-end text recognition using homogeneous and heterogeneous target datasets respectively. 

The potential of the FedTR framework is not only limited to text recognition applications. FedTR is adaptable and can be a framework template for many applications, not limited to applications in the manufacturing industry. With the users’ privacy well preserved and good performance, FedTR can be easily deployed to other deep learning applications. Our work has been practically implemented in the factory and applied in the digital manufacturing pipeline for the purpose of automated quality inspection




\bibliographystyle{ACM-Reference-Format}
\bibliography{sample-base}


\end{document}